\documentclass{article}

\usepackage{arxiv}

\usepackage[utf8]{inputenc} 
\usepackage[T1]{fontenc}    
\usepackage{hyperref}       
\usepackage{url}            
\usepackage{booktabs}       
\usepackage{amsfonts}       
\usepackage{nicefrac}       
\usepackage{microtype}      
\usepackage{lipsum}

\title{Causal Adversarial Network for Learning Conditional and Interventional Distributions}

\author{
  Raha Moraffah \\
  Department of Computer Science\\
  Arizona State University\\
  Tempe, AZ 85281 \\
  \texttt{Raha.Moraffah@asu.edu} \\
   \And
   Bahman Moraffah \\
  School of Electrical, Computer, \\and Energy Engineering \\
Arizona state university \\
Tempe AZ, 85287\\
  \texttt{Bahman.Moraffah@asu.edu} \\
   \And
   Mansooreh Karami \\
  Department of Computer Science\\
  Arizona State University\\
  Tempe, AZ 85281 \\
  \texttt{mkarami@asu.edu} \\
   \And
   Adrienne Raglin  \\
  Army Research Lab\\
  2800 Powder Mill Rd\\
  Adelphi, MD 20783 \\
  \texttt{adrienne.raglin2.civ@mail.mil } \\
   \And
 Huan Liu\\
  Department of Computer Science\\
  Arizona State University\\
  Tempe, AZ 85281 \\
  \texttt{hliu@asu.edu} \\
}

\newcommand{\raha}[1]{\textcolor{red}{{#1}}}
\newcommand{\m}{{CAN}}
\newcommand{\Gone}{{Label Generation Network}}
\newcommand{\Gtwo}{{Conditional Image Generation Network}}
\newcommand{\abbgone}{{LGN}}
\newcommand{\abbgtwo}{{CIGN}}
\usepackage{tikz}
\usetikzlibrary{bayesnet}
\usetikzlibrary{arrows}
\usepackage{color}
\usepackage{graphicx}
\usepackage{amsmath}
\usetikzlibrary{backgrounds}
\DeclareMathOperator{\E}{\mathbb{E}}
\newcommand{\norm}[1]{\left\lVert#1\right\rVert}
\usepackage{siunitx}
\usepackage{subfig}
\usepackage{multirow}
\usepackage{float}

\begin{document}

\maketitle

\begin{abstract}
We propose a generative Causal Adversarial Network (CAN) for learning and sampling from conditional and interventional distributions.  In contrast to the existing CausalGAN which requires the causal graph to be given, our proposed framework learns the causal relations from the data and generates samples accordingly. 
The proposed {\m} comprises a two-fold process namely  {\Gone} (\abbgone) and {\Gtwo} (\abbgtwo). The {\abbgone} is a GAN-based architecture which learns and samples from the causal model over labels. The sampled labels are then fed to {\abbgtwo}, a conditional GAN architecture, which learns the relationships amongst labels and pixels and pixels themselves and generates samples based on them. This framework is equipped with an intervention mechanism which enables the model to generate samples from interventional distributions. We quantitatively and qualitatively assess the performance of {\m} and empirically show that our model is able to generate both interventional and conditional samples without having access to the causal graph for the application of face generation on CelebA data.
 


\end{abstract}

\section{Introduction}
\label{introduction}

Generative Adversarial Networks (GANs) \cite{goodfellow2014generative} are ubiquitous tools for non-parametric sampling from complicated and high-dimensional distributions. GANs have achieved promising results in generating sharp-looking and realistic images and videos \cite{brock2018large, tulyakov2018mocogan}. They are also exploited to generate samples from categorical distributions \cite{camino2018generating} as well as text data \cite{yu2017seqgan}. One well-known extension of GAN is conditional GAN (cGAN) which enables sampling from conditional distributions by providing additional information such as class labels to both generator and discriminator. 
Recently, several cGAN frameworks have been developed for different purposes such as class conditional image generation \cite{mirza2014conditional, odena2017conditional}, generating image from text \cite{reed2016generative, zhu2019dm} and image to image translation \cite{isola2017image, zhu2017unpaired}. In this paper, we are focusing on class conditional image generation task for the case where multiple labels are conditioned on. As mentioned earlier, several class-conditional GAN models with impressive performances have been proposed \cite{mirza2014conditional, odena2017conditional}.
~However, all these frameworks do not model the relationships between the labels. This results in the model being incapable of conditioning on one or a set of labels and generating the remaining labels and images given the chosen label(s). In such cases, traditional cGANs sample the labels independently which means choosing the value of one label does not change the distribution of others. However, generating labels this way may lead to unexpected and unrealistic samples. For example, consider the scenario in which a cGAN is trained to generate person's face images conditioned on two class labels ``gender" and ``mustache". Class labels ``gender" and ``mustache" are obviously not independent. However, using traditional cGAN, if one conditions on the ``mustache = true" and samples ``gender" independently, the model is likely to generate images of females with mustache. Whereas in reality, we expect that conditioning on the ``mustache" affects the distribution of ``gender" as well and the model only generates images of males with mustaches. This is possible if we know the causal relationships between the labels and how they affect each other. 



Moreover, questions such as \textit{``what if the person in the image had mustache or was bald?"} are interesting questions and could lead into generating interesting samples that do not belong to the observed data. In order to model causal relations and answer ``what if?" questions, causal inference provides powerful tools, i.e. Structural Causal Models (SCMs), as a way of encoding causal relationships, and intervention mechanisms \cite{10.5555/1642718} as a tool to answer ``what if?" questions. Intervention on one variable is different from conditioning on it in the sense that the latter affects the distributions of both ancestors and descendants of the variable in the causal graph whereas the intervention only affects the distribution of descendants. For instance, In the aforementioned example, suppose we have ``gender causes mustache" ($gender \rightarrow mustache$) as a part of the causal graph, conditioning on mustache changes the distribution of the gender and thus, we only see males with mustache in the generated samples. Intervening on mustache on the other hand does not affect the gender and hence, we expect to see both males and females with mustache in our samples. Note that this is different from traditional cGAN's approach, since intervening on the variable in a causal graph affects the distribution of its descendants whereas choosing labels independently discards the relationships between the labels and does not affect the distribution of other labels at all. the distribution resulted from intervening on some variables is called the interventional distribution.
To consider dependencies between labels and enable cGANs to allow conditioning/intervening on a set of labels and generate the rest and the image accordingly, CausalGAN \cite{kocaoglu2017causalgan} proposes an adversarial training framework to learn a causal generative model based on a given causal graph. Despite promising results, this framework requires the causal graph of the labels to be known which cannot be satisfied in most real-world cases. Moreover, since the label generator in CausalGAN contains one neural net per each node in the causal graph of the labels, the model is not scalable to the large number of labels.

To address these problems, we propose a novel and scalable generative \textbf{C}ausal \textbf{A}dvesarial \textbf{N}etwork ({\m})  which aims to learn the causal relationships from the data.
{\m} is a 2-fold framework which consists of a {\Gone} ({\abbgone}) and {\Gtwo} ({\abbgtwo}). The {\abbgone} is trained to learn the causal model over the labels from the data and generates samples from the learned model. The labels are then fed to the {\abbgtwo}, an extension of AC-GAN \cite{odena2017conditional} which is designed to take in a set of labels, learn the label-pixel and pixel-pixel relationships from the data and generate the images accordingly.
To enable the {\m} to sample from interventional distributions, an intervention mechanism is proposed.

Our contributions summarized as follows: 1) We propose a novel architecture called {\m}, which generates high quality and diverse samples from conditional and interventional distributions without requiring the causal graph to be known; 2) We propose a novel intervention mechanism for {\m} which enables generating samples from interventional distributions using cGAN; and 3) We perform extensive experiments to assess the effectiveness of our model.

\section{Causal Adversarial Network ({\m})}\label{mainframework}
In this section, we introduce generative \textbf{C}ausal \textbf{A}dversarial \textbf{N}etwork ({\m}), an extension of class conditional GAN, which enables sampling from both conditional distributions (where some class labels are conditioned on and rest of the labels and the image are sampled given these conditions) and interventional distributions (where some class labels are intervened on and the remaining labels and the image are sampled accordingly) without requiring the causal graph over the labels to be known. Our model is specifically designed for the case where multiple categorical class labels are available and conditioned on per each image.
{\m} consists of a {\Gone} ({\abbgone}) and a {\Gtwo} ({\abbgtwo}). The {\abbgone} is a GAN-based framework which learns the causal relationships amongst the labels from the data and samples from it given a noise input. The sampled labels along with a random noise are then fed to the {\abbgtwo}, a novel extension of cGAN which learns the causal relations between labels and concepts in the image and concepts themselves and generates samples accordingly.
In order to learn the causal relations, we propose to modify the GAN's generator by integrating the Structural Causal Model (SCM) of the data into the generator and generating samples from it. The parameters of the SCM, the generator and discriminator are learned via adversarial training. While sampling from conditional distributions using {\m} is straightforward, sampling from interventional distributions is more complicated. To enable {\m} to generate samples from interventional distributions as well, we propose an intervention mechanism for the framework.


\subsection{Proposed framework} 
In this section, we demonstrate the general idea of embedding the SCM into the GAN's architecture using one of the widely used variant of GAN, WGAN-GP \cite{gulrajani2017improved} and introduce an intervention mechanism for this framework. We then describe how this idea contributes to components of {\m}.

\subsubsection{WGAN-GP}
A Generative Adversarial Network (GAN) \cite{goodfellow2014generative} consists of two neural networks namely generator and discriminator (critic) competing against each other. The generator takes in a random noise vector $z$ and generates a fake sample. The discriminator receives a real or a fake sample and determines whether it is synthesized by the generator or drawn from the real distribution. Different loss metrics have been proposed for GANs. For instance, Wasserstein-GAN (WGAN) \cite{arjovsky2017wasserstein} uses Earth-Mover (a.k.a Wasserstein-1) distance to compare the real and generated distributions. 
This metric is specifically suitable due to its convergence properties and correlation with the perceptual quality of generated images. In this paper, we adopt an improved and more stable version of WGAN, WGAN-GP, which enforces the Lipschitz constraint required by the objective of WGAN by adding a gradient penalty term. The discriminator and generator losses of WGAN-GP are respectively defined as:
\begin{equation}
\begin{aligned}
\label{EQ:WGAN-d}
& \mathcal{L}_{D} = \underbrace{\underset{\tilde{x} \sim
 \mathbb{P}_g }{\E} [D(\tilde{x})] - \underset{x \sim
 \mathbb{P}_r }{\E} [D(x)]}_{\text{Critic loss}} + \underbrace{\lambda~  \underset{\hat{x} \sim \mathbb{P}_{\hat{x}}}{\E} [(\norm{\nabla_{\hat{x}} D(\hat{x}) -1})^2]}_{\text{gradient penalty}},\\
 & 
\mathcal{L}_{G} = -\underset{\tilde{x} \sim
 \mathbb{P}_g }{\E} [D(\tilde{x})],\\
\end{aligned}
\end{equation}

where D is a set of 1-Lipschitz functions, $\mathbb{P}_r$ and $\mathbb{P}_g$ denote the real and model distributions, and $\mathbb{P}_{\hat{x}}$ is a distribution obtained by randomly interpolating between real and generated images.
The generator learned via this objective is a deterministic transformation from an easy-to-sample independent distribution to the target random variable. Once the parameters are learned, the generator can be used to perform non-parametric sampling from marginal distributions $P(x)$ with this generative process $\tilde{x} = G(z; \theta_g)~, z\sim p(z)$. The $G(z; \theta_g)$ is the generator parameterized with a neural network with parameters $\theta_g$ and trained via adversarial training.
~The graphical model of this process can be described as $z \rightarrow x$ which assumes that every variable in $x$ depends on every variable of $z$ \cite{DBLP:journals/corr/Goodfellow17}. However, in reality, this may not be true. To discover the relationships between variables in z and x and relations between variables in x from the data and generate samples based on them, we propose to embed a Structural Causal Model (SCM) in the generator's structure, learn its parameters along with other parameters of the model and generate samples using ancestral sampling.
\subsubsection{Integrating SCM into the GAN's Generator} \label{integration}
In this section, we explain the process of integrating the SCM into the GAN's generator by first formally introducing the SCM and then explaining the process of embedding it into GAN. 

Let $ X \in \mathbb{R} ^{n}$ be a sample from the joint distribution of n variables and $A \in \mathbb{R} ^{n \times n}$ be a weighted adjacency matrix of the causal DAG in which each node corresponds to a variable in X. The linear Structural Causal Model (SCM) is defined as:
\begin{equation*} \label{linear SEM}
 X = A^T X + Z
\end{equation*}
where $Z \in \mathbb{R} ^{n}$ is a random noise. In other words, in a SCM, each variable is defined as a (linear) function of its parents in the causal DAG and a noise variable. The child-parent relationships are encoded via the (weighted) adjacency matrix $A$. If the causal DAG is sorted in topological order, ancestral sampling can be done by first sampling a random noise Z and then solving the following triangular system:
\begin{equation*} \label{samplingtriangular}
 X = (I - A^T)^{-1} Z
\end{equation*}
This equation is a linear deterministic transformation from $Z$ to $X$. GAN generators are often non-linear transformations of noise. To add non-linearity, inspired by recent work on learning non-linear SCM \cite{lachapelle2019gradient, yu2019dag}, we propose to replace traditional generator ($G(Z; \theta_g)$) with the following:
\begin{equation} \label{samplingtriangular2}
 X = G((I - A^T)^{-1} Z; \theta_g, A)
\end{equation}
  where G is the generator in the traditional GAN and is instantiated based on the type of data. In this paper, we call G the non-linear transformation function. Equation \eqref{samplingtriangular2} demonstrates a mapping from noise to data space by taking causal relations into account. The parameter A is simultaneously learned along with all other parameters of the model via adversarial training.


Optimizing the adversarial loss does not guarantee the A to be a DAG as required by SCMs. A graph needs to satisfy the acyclicity constraint to be a DAG. To satisfy this condition, we impose an equality constraint which guarantees that a graph is acyclic if and only if for any $\beta > 0$ \cite{yu2019dag}:
\begin{equation} \label{acyclicity}
 tr[(I + \beta A \odot A)^n] - n = 0
\end{equation}
where $A \in \mathbb{R} ^{n \times n}$ is the weighted adjacency matrix of the graph, n is number of nodes, ``tr" represents trace of a matrix and $\odot$ is element-wise multiplication.
We therefore combine this equality constraint with our adversarial loss. The objectives of the modified WGAN-GP with embedded SCM are given as:
\begin{equation}
\begin{aligned}
\label{EQ:main1-d}
& \mathcal{L}_{D} = \underset{\tilde{x} \sim
 \mathbb{P}_g }{\E} [D(\tilde{x})] - \underset{x \sim
 \mathbb{P}_r }{\E} [D(x)] + \lambda~  \underset{\hat{x} \sim \mathbb{P}_{\hat{x}}}{\E} [(\norm{\nabla_{\hat{x}} D(\hat{x}) -1})^2],\\
 & 
\mathcal{L}_{G} = -\underset{\tilde{x} \sim
 \mathbb{P}_g }{\E} [D(\tilde{x})] = -\underset{z \sim
 \mathbb{P}_z }{\E} [D(G(z;\theta, A))]\\
& \text{s.t.}~~~tr[(I + \beta A \odot A)^n] - n = 0
\end{aligned}
\end{equation}
This objective is solved with augmented Lagrangian approach \cite{bertsekas1999nonlinear}. More details can be found in the Appendix. 

Note that recovering causal graphs from observational data without further assumptions on data generation process (such as additive Gaussian noise assumption) is generally impossible \cite{peters2012identifiability,peters2014causal}. Since our framework is designed for real world data which often do not satisfy such assumptions, we cannot theoretically guarantee whether the true causal graph is identifiable or not. However, Our empirical results demonstrate the effectiveness of our proposed model in learning the causal graph from the observational data.

\subsubsection{Intervention Mechanism for Sampling from Interventional Distributions}

In a SCM, each variable $x_i$ can be written as a deterministic function of its parents ($pa(x_i)$) and a noise variable ($z_i$). Intervention on variable $x_i = f_i(pa(x_i), z_i)$ in the system is accomplished by replacing the function $f_i$ with the desired value  where the rest of the system remain unchanged \cite{10.5555/1642718}. In terms of graphical models, this is equivalent to removing all incoming edges to the node in the causal DAG and replacing the value of that node with the interventional value. Doing so makes the distribution of the ancestors of the node unchanged and only changes the distribution of the descendants. To model this process, we consider two sources of input for each node in the causal DAG: 1)values of the parent nodes and a noise variable and 2)interventional value for that node. In the case of generating purely observational data, the values of parents and a random noise are sampled and used to calculate the value of the node using the equation $x_i = f_i(pa(x_i), z_i)$. Whereas, in the case of sampling from interventional distribution, the interventional value is selected as the value of the node and that value is propagated to the descendants of that node in the graph. This selection process is implemented via a mask vector $\alpha_1$ which performs as selector and for each node either selects the value of parents, their corresponding weights in the SCM and a random noise and calculates the final value the node using them or selects the interventional value and sets it as the value of node. More formally, we extend our mapping function defined in equation \eqref{samplingtriangular2} to enable sampling from both conditional and interventional distributions as follows:
 \begin{equation} \label{interventionalsampling}
 X = G((I - \alpha_1 \odot A^T)^{-1} (\alpha_1 \odot Z + (1 - \alpha_1) \odot C))
\end{equation}
where $C \in \mathbb{R} ^{n}$ is a vector of interventional values for nodes in the causal DAG where $C_i$ corresponds to the desired interventional value for the $i-th$  node of the graph and $\alpha_1 \in \mathbb{R} ^{n}$ is a selector mask which selects source of inputs for each node in the graph. If $\alpha_{1i} = 0$ then the equation for $x_i$ is reduced to equation \eqref{samplingtriangular2} which basically means the value of $x_i$ is calculated as a nonlinear function of its parents and a random noise and if $\alpha_{1i} = 1$ then $x_i$ is forced to be equal to $C_i$, i.e., $x_i = C_i$ which will be propagated and used in calculating all of its descendants. Note that since $\alpha_1$ is a vector, to be able to perform its element-wise multiplication with a matrix (A), during the implementation, we broadcast $\alpha_1$ so that they have the same shape.
 
Note that we use the intervention mechanism to sample from interventional distribution at inference time when the SCM has already been learned. 
Note that generating samples from conditional distributions using this framework can be performed via rejection sampling.

Next, we discuss each component of the {\m} framework and how the method proposed in the previous section can be used to model them.

\subsubsection{{\Gone}}
\label{Framework:G1}

We introduce the {\Gone} ({\abbgone}) whose aim is to learn the causal model of the labels and generate samples based on them. To achieve this, We directly use the generator proposed in equation \eqref{samplingtriangular2}. However, the generator's non-linear transformation function (G), the discriminator (D) and the loss criterion still need to be designed according to the type of the data the GAN is designed to generate. 

Labels in commonly used datasets are often of discrete nature with multiple categorical variables. Therefore the choice of $G$, $D$ and the objective loss need to be suitable for this data type. 
For the G, we employ the architecture proposed in \cite{camino2018generating} for multi categorical data. The architecture consists of multiple fully connected layers followed by a layer with one fully connected output layer per each categorical variable (i.e. label) in the sample. A softmax activation is then applied on top of this layer. Finally, the outputs of the softmax are concatenated with each other and generate the final output of the generator. The discriminator is a set of fully connected layers. 
For the adversarial loss, we utilize WGAN-GP's loss, which has proven to be capable of generating discrete samples using continuous generators \cite{gulrajani2017improved}. We learn the parameters of {\abbgone} by optimizing the adversarial loss in equation \eqref{EQ:main1-d}.  
\subsubsection{{\Gtwo}}
\label{Framework:G2}
Even though the generator should be able to converge eventually as long as it is capable of generating the joint distribution represented by the observations, it is shown that if the generator is structured according to the true underlying causal graph, it is expected to converge faster (within a fewer training steps) \cite{kocaoglu2017causalgan}. Motivated by this, 
we propose a novel cGAN architecture called {\Gtwo} ({\abbgtwo}), which models the causal relations between the labels and pixels and amongst pixels themselves. 

In traditional cGANs, labels and random noise are simply concatenated and fed to the generator to synthesize fake samples. These models assume all labels affect all pixels in the image which may not be true in reality. To learn the relationships between label and pixel and pixels themselves and consider them while generating samples, we propose a new architecture for cGAN's generator. While learning pixel-pixel relations can be accomplished by directly using the framework in equation \eqref{EQ:main1-d}, that model is specifically designed for sampling from marginal distributions and does not support conditioning on additional information (labels). 
To enable conditioning on labels, we leverage our proposed intervention mechanism. Despite differences between conditioning and intervening on a variable, i.e, conditioning affects the distributions of both ancestors and descendants and intervening only affects the descendants in the causal graph, these two concepts operate the same for the variables who have no ancestors in the graph. Since in image generation process the assumption is that labels cause the image, $L \rightarrow I$, the labels are the ancestors of the pixels of the image and conditioning on them is equivalent to intervention. This subtle observation allows us to use the generator in equation \eqref{EQ:main1-d} with minor modifications as cGAN generator. Particularly, our generator learns the causal model of labels and pixels, but the values of labels (generated by LGN)  are already set by intervening on them. Hence, since the values of labels are already given, we don't need to know the relationships between them and only need to learn the label-pixel and pixel-pixel relations. 

\begin{figure}
\centering
\includegraphics[height = 1.75in, width=90mm]{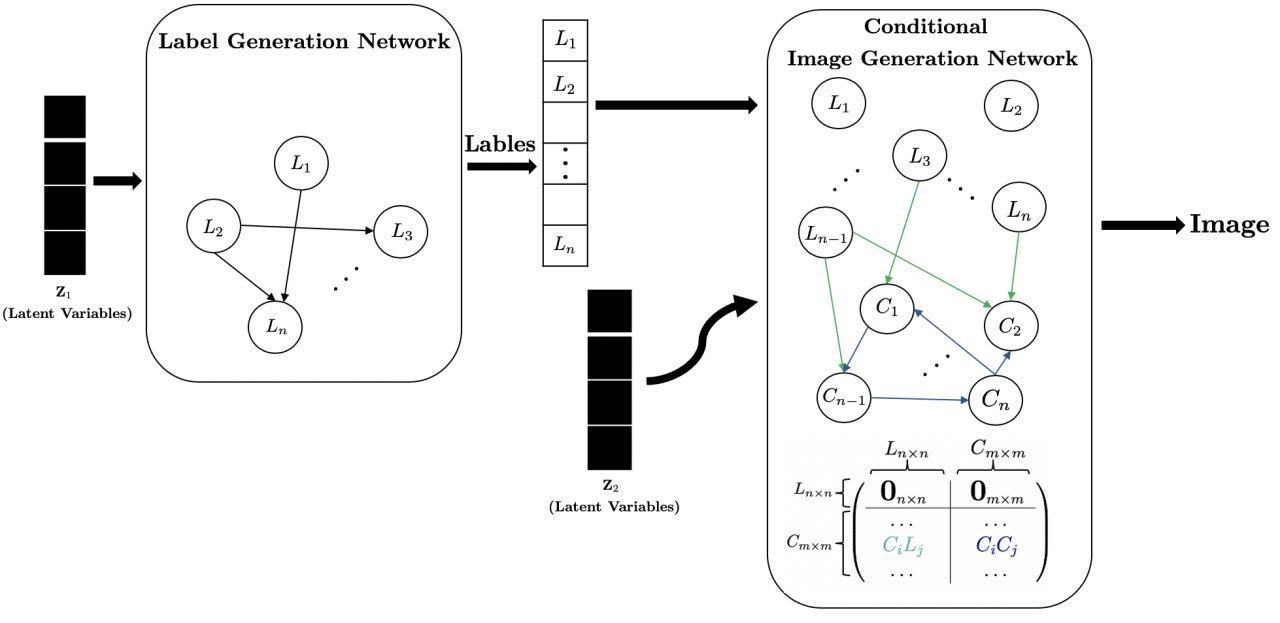}
\caption{{\m} framework at a glance: The framework consists of a {\abbgone} followed by {\abbgtwo}. The {\abbgone} learns the causal relations between labels and samples from them. The labels are fed to {\abbgtwo} which learns the label-pixel and pixel-pixel relations to generate samples conditioned on labels provided by{\abbgone} .} 
\label{fig:main}
\end{figure}

Now we discuss the details of the model.
For the choice of G, broadly speaking, the network comprises a series of ‘deconvolution’ layers and the discriminator network is a series of 'convolutional' layers.
In order to learn the parameters, we modify a variation of AC-GAN's objective based on WGAN-GP loss to support multiple labels. Formally the objectives of our proposed {\abbgtwo} are defined as:
\begin{equation}
\begin{aligned}
\label{EQ:cgan}
& \mathcal{L}_{D} = \underset{\tilde{x} \sim
 \mathbb{P}_g }{\E} [D(\tilde{x})] - \underset{x \sim
 \mathbb{P}_r }{\E} [D(x)] + \\
& \lambda~  \underset{\hat{x} \sim \mathbb{P}_{\hat{x}}}{\E} [(\norm{\nabla_{\hat{x}} D(\hat{x}) -1})^2]- \mathcal{L}_{C},\\
&\mathcal{L}_{G} = -\underset{\tilde{x} \sim
 \mathbb{P}_g }{\E} [D(\tilde{x})] - \mathcal{L}_{C} = -\underset{z \sim
 \mathbb{P}_z }{\E} [D(G(z, C;\theta, A))] - \mathcal{L}_{C}\\
& \text{s.t.}~~~tr[(I + \beta A \odot A)^n] - n = 0
\end{aligned}
\end{equation}
where $ \mathcal{L}_{C}$ is the log-likelihood of the correct class defined as $\mathcal{L}_{C} = {\E} [\log P(C = c~|X_{\text{real}})] + {\E} [\log P(C = c~|X_{\text{fake}})]$.
These objectives are solved via adversarial training. More details on the architecture of both {\abbgone} and {\abbgtwo} are discussed in the Appendix. Finally, the {\abbgone} and {\abbgtwo} combined together are called {\m}. The labels are first sampled from their conditional or interventional distributions via {\abbgone}. The sampled labels are then fed to the {\abbgtwo} and generate samples given the labels. Figure \ref{fig:main} displays an overview of our proposed {\m} framework.

\section{Experiment}\label{experiment}

In this section, we evaluate the performance of the proposed {\m} from both quantitative and qualitative perspectives. In our experiments, we aim to answer the following questions: 1) Is {\m} capable of generating samples from both conditional and interventional distributions? Are the generated images diverse, of high quality and close to the real distribution?; 2) Are labels generated by {\abbgone} component of {\m} high quality?; and 3) Is {\m} able to learn the causal graph from the data?
 ~To answer these, we perform three types of experiments: image generation evaluation,  label generation evaluation, and validation of the causal graph learned by the {\m}.
 More details on the experimental settings are explained in the Appendix.


\subsection{Image Generation Evaluation}
Here, we seek to answer the first experimental question, i.e., examine {\m}'s ability to generate samples from conditional and interventional distributions and assess the quality of the images generated by {\abbgtwo} component of CAN. Evaluating GAN's performance is a challenging task~\cite{theis2015note}.
  ~Following state-of-the-arts \cite{odena2017conditional, kocaoglu2017causalgan,brock2018large}, we assess the performance of our proposed framework through both qualitative and quantitative experiments. 
To demonstrate the effectiveness of {\m}, we train the model on CelebA \cite{liu2015deep}, a large-scale face attributes dataset, which includes 202,599 images of faces of celebrities along with 40 binary attribute annotations/labels per each image. Even though {\m} is capable of being trained on all 40 attributes, to make a fair comparison with previous work \cite{kocaoglu2017causalgan}, the results are presented on 9 selected labels, i.e., ``Bald","Eyeglasses","Male","Mouth-Slightly-Open","Mustache","Narrow-Eyes","Smiling","Wearing-Lipstick", and ``Young".

\textbf{Qualitative Evaluation}

\begin{figure*}%
    \centering
    \subfloat[]{{\includegraphics[width=8cm, height = 3.7cm]{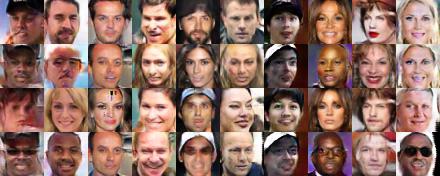} }}\hspace{0.7mm}%
    \qquad
    \subfloat[]{{\includegraphics[width=8cm, height = 3.7cm]{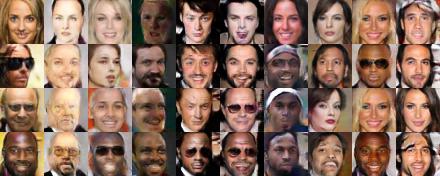} }}%
    \caption{Results of {\m} for intervening and conditioning on labels. Top two rows show the results for intervention (i.e. intervention on label=0 and label=1). Bottom two rows are the results for conditioning (i.e. condition on label=0 and label=1) : (a) shows the results for ``Bald" label.
    Since the causal graph is expected to be $Male \rightarrow Bald$, in the interventional samples we have both bald males and females. However, in the conditional samples, only bald males can be found. (b) shows the results on label ``Mustache". The results are in compliance with the expected causal graph $Male \rightarrow Mustache$. Therefore, in the interventional samples we have both males and females with mustache. However, in the conditional samples, only males with mustache can be found.}%
    \label{fig:bald}%
\end{figure*}

Here we verify if {\m} is able to generate samples from both interventional and conditional distributions. Samples from conditional distribution are generated by conditioning on a class label and generating the remaining labels as well as the image, given the specified label. Samples from interventional distribution are generated by intervening on an arbitrary label and generating the rest of labels and the image according to value of the given label. 
The difference between these two distributions are justified by adding/removing certain feature to the image through intervention or conditioning on the class labels and analyzing the resultant images. Figure \ref{fig:bald} illustrates the samples generated from both conditional and interventional distributions and their differences. Due to the space limitation, we only showcase our results for two labels, i.e., \textit{Mustache} and \textit{Bald}. The results for other labels are presented  in the Appendix.

\textbf{Quantitative Evaluation}

In this section, we quantitatively asses the quality of the images generated by {\abbgtwo} and compare our results with the following baselines: 1) AC-GAN \cite{odena2017conditional}: The state-of-the-art cGAN designed for class-conditional image synthesis. 
In AC-GAN the generator is provided with a noise vector and a class label and the discriminator is designed to predict the class label as well as the image authenticity.
AC-GAN does not model the label-label relationships and do not learn the label-pixel and pixel-pixel causal relations. In our experiments we implement the AC-GAN with WGAN-GP loss and also extend the architecture to accept multiple labels; 2) CausalGAN \cite{kocaoglu2017causalgan}: The state-of-the-art GAN designed for sampling from both interventional and conditional distributions based on a known causal graph which is provided by authors of the paper. CausalGAN only considers causal dependencies between the labels and does not consider the causal relations between the labels-pixel and pixel-pixel. The main limitation of causalGAN is that it requires the causal graph between the labels to be known. 
Note that both baselines and {\m} are trained for same number of epochs.

In order to gauge the visual quality of synthesized images, we calculate Fréchet Inception Distance (FID) which measures the difference between the distributions of generated and real images activations when fed to the Inception network \cite{heusel2017gans}. FID score is  specifically designed to evaluate the quality of images sampled from the marginal distributions and is not tailored for cGANs which generate samples from conditional distributions. Since our method can be considered as an extension of cGAN, we also report GAN-train and GAN-test metrics \cite{shmelkov2018good} which are particularly proposed to evaluate the performance of cGANs. These two metrics are based on classification accuracy and approximate the recall (diversity) and precision (quality of the image) of cGANs, respectively. Since these metrics are originally proposed for the case where each image exactly has one label, to be able to apply them to our case, we extend them by using Hamming score (a.k.a. label-based accuracy) which is widely used for multi-label classification instead of accuracy. We also report the GAN-train and GAN-test for each label individually along with details on these metrics calculations in the Appendix. 
Table \ref{table1:metrics} illustrates the FID, GAN-train and GAN-test scores achieved by our model (denoted by {\abbgtwo}-{\m}) and the baselines. Our results show that CAN outperforms both baselines. This demonstrates the effectiveness of the proposed framework in generating high quality and diverse images in addition to being capable of generating samples from both interventional and conditional distributions and also suggests that learning  causal relations  helps improving the quality of generated images considering that all models are trained for the same number of epochs. 



\begin{table}[h!]
  \begin{center}
  \caption{The FID (lower is better), GAN-train and GAN-test scores (higher is better) comparisons on CelebA data}
    \label{table1:metrics}
    \begin{tabular}{@{}cccc@{}} \toprule

Model &  FID & GAN-train & GAN-test\\ \midrule
{\abbgtwo}-{\m} & 4.95& 0.65& 0.62\\
CausalGAN  & 20.32& 0.58 & 0.45\\
AC-GAN   & 12.58& 0.60& 0.42\\\bottomrule
\end{tabular}
  \end{center}
  
\end{table}




\subsection{Label Generation Evaluation}

\begin{table*}[t]
\centering
\caption{Evaluation of the quality of labels on Child dataset} 
\begin{tabular}{@{\extracolsep{4pt}}cccc}
\toprule   
 {} & \multicolumn{3}{c}{Child} \\
 \cmidrule{2-4} 

 Model  &  $\text{MSE}_p$ & $\text{MSE}_f$ & $\text{MSE}_a$ \\ 
\midrule
{\abbgone}-{\m} & \num{5.1e-4} $\pm$ \num{8e-5} & \num{1.4e-3} $\pm$ \num{4e-4} & \num{3.4e-4} $\pm$ \num{7e-5} \\ 
 
MC-WGAN-GP & \num{9.8e-4} $\pm$ \num{5e-5}& \num{1.8e-3} $\pm$ \num{4e-4}& \num{4.2e-4} $\pm$ \num{8e-5} \\ 

Causal Controller & - & - & - \\ 
\bottomrule
\end{tabular}

\label{table:childlabel}
\end{table*}

\begin{table*}[t]
\centering
\caption{Evaluation of the quality of labels on CelebA-label dataset} 
\begin{tabular}{@{\extracolsep{4pt}}cccc}
\toprule   
 {} & \multicolumn{3}{c}{CelebA-label} \\
 \cmidrule{2-4} 

 Model  &  $\text{MSE}_p$ & $\text{MSE}_f$ & $\text{MSE}_a$ \\ 
\midrule
{\abbgone}-{\m} & \num{6.9e-4} $\pm$ \num{e-5}  & \num{2.2e-4} $\pm$ \num{5e-5} & \num{1.4e-5} $\pm$ \num{3e-5} \\ 
 
MC-WGAN-GP& \num{6e-4} $\pm$ \num{e-5}& \num{6e-4} $\pm$ \num{e-4}& \num{1.3e-5} $\pm$ \num{3e-5}\\ 

Causal Controller & \num{7.2e-4} $\pm$ \num{e-5}& \num{3.2e-4} $\pm$ \num{3e-4}& \num{1.1e-4} $\pm$ \num{e-5} \\ 
\bottomrule
\end{tabular}

\label{table:celeblabel}
\end{table*}

In this section, we evaluate the quality of generated labels by assessing
the performance of {\abbgone} component of CAN in generating multi categorical data. 
We compare the \abbgone ~with two baselines:1)  MC-WGAN-GP \cite{camino2018generating}, the state of the art GAN for synthesizing multi categorical data, which does not model causal relations; and 2) Causal Controller, a component of CausalGAN which is in charge of learning and sampling from labels by modeling each function in the SCM corresponding to the given causal graph of the labels with a feed forward neural network. Since in the Causal Controller, there is one NN per each label, the model cannot be scaled to high number of labels.
We demonstrate our results on three datasets with multiple categorical variables: Child and Alarm datasets \cite{tsamardinos2006max} and All 9 labels from CelebA data used in the previous experiment (CelebA-label data). These datasets are specifically chosen to measure the effectiveness of the models in generating multi categorical samples as well as their scalability to high number of labels.
The description of these datasets can be found in the Appendix.
In our experiments, we split the data into 90\% training and 10\% test. We utilize three Mean Squared Error (MSE) based metrics, i.e. $\text{MSE}_p = \text{MSE}(p_{\text{test}}, p_{\text{sample}})$, $\text{MSE}_f = \text{MSE}(f_{\text{train}}, f_{\text{sample}})$ and $\text{MSE}_a = \text{MSE}(a_{\text{train}}, a_{\text{sample}})$ used in \cite{choi2017generating, camino2018generating} to evaluate the quality of generated multi categorical samples. In these metrics, $p_{\text{test}}$ and $p_{\text{sample}}$ are vectors of frequencies of ones corresponding to real samples in the test and generated samples, respectively. $f_{\text{train}}$ and $f_{\text{sample}}$ are the f-1 scores of prediction of each dimension of the vector representing a sample using a logistic regression model trained on both real train set and generated samples.  $a_{\text{train}}$ and $a_{\text{sample}}$ are the accuracy of predicting each categorical variable in the multi categorical sample using real train and generated samples.
We report the results for Child and CelebA-label data in tables \ref{table:childlabel} and \ref{table:celeblabel} and the results for Alarm can be found in the Appendix. Note that Causal Controller cannot be trained on child dataset (with 20 categorical variables per each sample) due to scalability issue and therefore the results for this model is not reported.
Our results indicate that our model (shown with {\abbgone}-{\m}) outperforms the baselines in most cases which demonstrates the effectiveness of {\abbgone} in generating high quality labels. Comparing the results of {\abbgone} with MC-WGAN-GP indicates that considering causal relations can improve the quality of generated labels. For the CausalGAN, one potential reason for being outperformed by {\abbgone} is that the causal graph which the Causal Controller is built upon may not be completely correct.
\subsection{Validating the causal graph learned by the model}
Since the performance of the {\m} highly depends on the quality of the causal graphs it learns, in this section we examine the ability of the algorithm used in {\m} to discover the causal graph from data. We compare our method with both traditional causal discovery algorithms such as PC algorithm \cite{spirtes2000causation} and GSF \cite{huang2018generalized} and the recent gradient based algorithms such as DAG-GNN \cite{yu2019dag} and Causal discovery based on reinforcement learning, a.k.a. RL-BIC, \cite{zhu2019causal}. We perform our experiments on two commonly used datasets in causal discovery, i.e., Child and Alarm datasets \cite{tsamardinos2006max}. Since these datasets are discrete, we use {\abbgone} to learn the causal graph. We will describe these algorithms and their implementations in the Appendix. To assess the quality of the estimated graph, we utilize two metrics: True Positive Rate (TPR), for which higher values are better and structural Hamming distance (SHD) which is the smallest number of edge addition, deletion and reversal to convert the inferred graph into the groundtruth graph. The lower SHD means the estimated graph is closer to the groundtruth and therefore is better. As shown in table \ref{table2:resultschildalarm}, our model (shown with {\abbgone}-{\m}) outperforms the baseline in most of the cases, which illustrates the capability of {\abbgone} in learning the causal graph from the data.

We also provide the causal graph learned by {\abbgone} over CelebA-labels in the Appendix.
\begin{table}[ht]
\caption{Results on child and alarm dataset: SHD (the lower the better) and TPR (The higher the better)} 
\centering
\begin{tabular}{@{\extracolsep{4pt}}ccccc}
\toprule   
 {} & \multicolumn{2}{c}{Child}  & \multicolumn{2}{c}{Alarm}\\
 \cmidrule{2-3} 
 \cmidrule{4-5} 
 Model  &  SHD & TPR  & SHD & TPR \\ 
\midrule
{\abbgone}-{\m}  & 18 & 0.78 & 33 & 0.75 \\ 
 
DAG-GNN &19  & 0.68 & 43 & 0.62  \\ 

RL-BIC &18  & 0.68 & 30 & 0.80  \\ 

PC &22  & 0.72 & 31 & 0.67  \\ 

GSF &23  & 0.76 & 49 & 0.76  \\ 
\bottomrule
\end{tabular}

\label{table2:resultschildalarm}
\end{table}


\section{Related work}
\label{related}
Conditional GAN is a type of GAN which allows conditional image generation. 
Class conditional GAN is a type of cGan which generates images conditioned on class labels. Mirza and Osindero propose the first class conditional GAN in which labels of the images are fed to both generator and discriminator \cite{mirza2014conditional}.
Odena et al. \cite{odena2017conditional}  propose AC-GAN to improve the quality of the generated images and generate high resolution samples. This is achieved by modifying the discriminator to predict the class label of the images as well as their authenticity. 
between the labels and the images. 
In another attempt to combine labels with images, 
authors propose to learn the joint distribution of labels and images \cite{liu2016coupled, pu2018jointgan}.
However, aforementioned frameworks do not model the causal dependencies between the labels and hence do not allow conditioning/intervening on a set of labels and generating the rest of labels and the image accordingly. To extend traditional cGANs to have this functionality, kocaoglu et al. \cite{kocaoglu2017causalgan} proposes CausalGAN, an adversarial training procedure to generate conditional images given a causal graph for the labels. Despite considering the causal relations between the labels, the performance of CausalGAN is limited by the requirement of knowing the causal graph.

Causal discovery is the task of identifying the causal relationships between a set of variables. Constraint-based causal discovery methods find the causal skeleton using conditional independence tests and orient the edges up to the Markov equivalence class.
~PC algorithm \cite{spirtes2000causation} is one of the well known methods in this category. 
Score-based algorithms are another type of causal discovery methods which leverage a score function to measure how well a graph fits the data and utilize a search algorithm to explore the space of possible structures to find the best graph with the optimal score \cite{heckerman1999bayesian}. Due to the intractability of the search space, these methods usually impose additional assumptions to narrow the scenarios for which the methods are applicable.
For instance, GES \cite{chickering2002optimal} assumes a linear parametric model with Gaussian noise and searches the space of CPDAGS with a greedy algorithm to optimize the Bayesian Information Criterion. GSF \cite{huang2018generalized} adopts the same search strategy as GES but with a generalized score function which does not assume particular model classes. This enables GSF to model nonlinear causal relations for a wide class of data distributions. 
Recently, NOTEARS \cite{zheng2018dags} has proposed a continuous constrained optimization problem to find the optimal DAG. The problem can be solved with existing blackbox solvers and this way NOTEARS avoids the combinatorial constraints. DAG-GNN \cite{yu2019dag} extends NOTEARS to support nonlinear relationships by proposing a graph neural network based model to recover the causal DAG.
GraN-DAG \cite{lachapelle2019gradient} proposes a score-based method which extends continuous constrained optimization framework with neural network to discover nonlinear relationships between variables.
Zhu et al. \cite{zhu2019causal} propose to use reinforcement learning to search for the best fitting DAG in score-based causal discovery. Their proposed reward function consists of a score function and two penalty terms which account for acyclicity constraint. 

In recent years, causality has been also used to improve the deep neural nets. Besserve et al. explore the connection between GANs and causal generative models by considering the image as the cause of neural network's weights \cite{besserve2017group}. Lopez-Paz et al. \cite{lopez2017discovering} propose a neural net based approach to discover the causal relations between a label and a static image by discovering the causal directions. Odena et al. \cite{ odena2018generator} demonstrate that there exist causal relations between conditioning of the Jacobian and quantitative metrics for evaluating GAN. However, none of these works utilize causality to generate interventional and conditional images.
\section{Conclusions}
In this paper, we propose a generative Causal Adversarial Network (CAN) which learns the label-label, label-pixel and pixel-pixel causal relations from the data and generates samples accordingly. The framework consists of a {\abbgone} and {\abbgtwo}. The labels sampled from {\abbgone} are fed to the {\abbgtwo} and together they can be used to generate samples from conditional and interventional distributions. We validate the performance of our model through comprehensive experiments.

\section*{Appendix}

\section*{A.1. Optimization Algorithm using Augmented Lagrangian Multiplier}
In order to solve the objective described in equation (4) in the main text, we leverage augmented Lagrangian approach \cite{bertsekas1999nonlinear} which is widely used to solve the nonlinear equality-constrained
problems. We define the Augmented Lagrangian for equation (4) as:
\begin{equation}
\begin{aligned}
\label{EQ:augmented}
& \mathcal{L}_{D} = \underset{\tilde{x} \sim
 \mathbb{P}_g }{\E} [D(\tilde{x})] - \underset{x \sim
 \mathbb{P}_r }{\E} [D(x)] + \lambda~  \underset{\hat{x} \sim \mathbb{P}_{\hat{x}}}{\E} [(\norm{\nabla_{\hat{x}} D(\hat{x}) -1})^2],\\
& 
\mathcal{L}_{G} = -\underset{z \sim
 \mathbb{P}_z }{\E} [D(G(z;\theta, A))] + \bar{\lambda} c(A)   + \frac{\rho}{2} \norm{c(A)}^2,
\end{aligned} 
\end{equation}
where $c(A) = tr[(I + \beta A \odot A)^n] - n$ is the equality constraint and $\bar{\lambda}$ and $\rho > 0$ are the Lagrange multiplier and the quadratic penalty weight, respectively. Note that we only impose the constraint when training the generator.
We alternate between optimizing the discriminator and the generator. To optimize the objectives in the aforementioned equation, we use stochastic optimization solvers such as ADAM optimizer \cite{kingma2014adam} or RMSprop \cite{Tieleman2012}.

We update $\bar{\lambda}$ in each step with the following rule:
\begin{equation*} \label{EQ:update}
  \bar{\lambda} = \bar{\lambda} + \rho c(A) 
\end{equation*}

\section{A.2. Implementation Details}
\subsection*{A.2.1. {\abbgone} and {\abbgtwo} Network Architectures}
In this section, we describe the details of the architectures of G (the nonlinear transformation function used in the generator) and D (discriminator) for the models trained on each dataset. 

For all experiments, we use ADAM \cite{kingma2014adam} with $\beta_1 = 0.5$
and $\beta_2 = 0.999$ to optimize the {\abbgone} network and RMSprop \cite{Tieleman2012} to optimize the {\abbgtwo}. The learning rates for both generator and discriminator in {\abbgone} are 0.001. The learning rates for the generator and discriminator of {\abbgtwo} are set to 0.001 and 0.0002, respectively.
All {\abbgone} networks are trained for 250 epochs and the {\abbgtwo} network is trained for 300 epochs. We set the hyperparameter $\lambda$ (gradient penalty coefficient) to 1.
It is worth mentioning that the dimension of the input noise variable to the generator (z) is not necessarily the same as the dimension of the data and could be much smaller. Moreover, in both generator and discriminator of {\abbgone} network, each categorical variable is represented as a one-hot encoded vector and each sample, a multi-categorical variable, consists of multiple one-hot encoded vectors concatenated with each other.
For the generators, we mostly use normal
rectified linear units (ReLU) as nonlinearity and for the discriminators, we utilize
leaky rectified linear units (LReLU) \cite{maas2013rectifier} with the negative slope 0.2.

In what follows, we thoroughly discuss the details of the architectures of the models trained on each dataset. 
Table \ref{table:notation} summarizes the notations we used to descibe the architectures of {\abbgone} and {\abbgtwo}.

\begin{table}[!htbp]
\centering
\caption{Notations used to describe the architectures of {\abbgone} and {\abbgtwo}} 

\begin{tabular}{@{\extracolsep{4pt}}cc}
\toprule   
 \textbf{Notation} &  \textbf{Definition}\\
 \cmidrule{0-0}
 \cmidrule{2-2} 
DECONV & Deconvolutional layer\\ 
 \midrule
CONV & Standard convolutional layer \\ 
 \midrule
FC & Fully-connected layer \\ 
 \midrule
N & Number of output channels\\ 
 \midrule
K &  Kernel size\\ 
 \midrule
S & Stride size\\ 
 \midrule
P & Padding size\\
\midrule
BN &  Batch normalization\\
\bottomrule
\end{tabular}

\label{table:notation}
\end{table}

\subsection*{A.2.2. CelebA Dataset}
\textbf{CelebA Data Preprocessing:} The CelebFaces Attributes dataset \cite{liu2015deep} is a dataset of 202,599 faces of celebrities annotated with 40 attributes. To be consistent with the previous work \cite{kocaoglu2017causalgan}, we select the following 9 attributes to perform our experiments:``Bald","Eyeglasses","Male","Mouth-Slightly-Open","Mustache","Narrow-Eyes","Smiling","Wearing-Lipstick", and ``Young". 
We preprocess the images by first cropping the $178\times218$ images to $178\times178$ and then resizing them to $64\times64$. 

\textbf {Network Architecture:}
Table \ref{table:celeblgn} and \ref{table:celebign} show the architectures of {\abbgone} and {\abbgtwo} trained on CelebA data. For the {\abbgtwo} Network, we input a 9 dimensional vector containing the class labels along with 128 noise variables, which together results in a 137 dimensional input. For the input to the {\abbgone}, we use 9 noise variables.

\begin{table}[!htbp]
\centering
\caption{{\abbgone} Network Architecture for CelebA} 

\begin{tabular}{@{\extracolsep{4pt}}cc}
\toprule   
 \textbf{Discriminator} D &  \textbf{Generator} G \\
 \cmidrule{0-0}
 \cmidrule{2-2} 
$\text{Input} \in \mathbb{R}^{18}$ & $\text{Input} \in \mathbb{R}^{9}$\\ 
 \midrule
FC-(N100), LReLU(0.2) & FC-(N100), ReLU \\ 
 \midrule
FC-(N100), LReLU(0.2)& FC-(N100), BN, ReLU \\ 
 \midrule
FC-(N100), LReLU(0.2) & FC-(N100), BN, ReLU \\ 
 \midrule
FC-(N1) & FC-(N100), BN, ReLU \\ 
 \midrule
{} & $9 \times$FC-(N2), BN, ReLU \\ 
 \midrule
{} & $9 \times$ Softmax\\ 
\bottomrule
\end{tabular}

\label{table:celeblgn}
\end{table}

\begin{table*}[!htbp]
\centering
\caption{{\abbgtwo} Network Architecture for CelebA} 
\begin{tabular}{@{\extracolsep{4pt}}cc}
\toprule   
 \textbf{Discriminator} D &  \textbf{Generator} G \\
 \cmidrule{0-0}
 \cmidrule{2-2} 
Input $64 \times 64$ Color image & $\text{Input} \in \mathbb{R}^{137}$\\ 
 \midrule
CONV-(N64, K$4\times4$, S2, P1), LReLU(0.2) & FC-(N1024), BN, ReLU \\ 
 \midrule
CONV-(N128, K$4\times4$, S2, P1), BN, LReLU(0.2)& FC-(N$128\times16\times16$), BN, ReLU \\ 
 \midrule
FC-(N1024), BN, LReLU(0.2) & DECONV-(N64, K$4\times4$, S2, P1), BN, ReLU\\ 
 \midrule
GAN-disc: FC-(N1) & \multirow{2}{*}{DECONV-(N3, K$4\times4$, S2, P1), Tanh}\\ 
Class-disc: FC-(N9)& \\
\bottomrule
\end{tabular}

\label{table:celebign}
\end{table*}

\subsection*{A.2.3. Child and Alarm Datasets}

~~~~\textbf{Child Dataset:} Child data  \cite{tsamardinos2006max} is a commonly used dataset in causal discovery. This dataset consists of 5000 multi-categorical samples where each sample comprises 20 categorical features. Each category is in the range of 1-5.

\textbf {Alarm Dataset:} A widely used multi-categorical data in causal discovery with 5000 samples. Each sample consists of 37 categorical features and each category ranges from 1 to 3.

\textbf {Network Architecture:}
Since both Child and Alarm datasets consist of multi-categorical samples, we train the {\abbgone} Network on these datasets. Table \ref{table:childlgn} and \ref{table:alarmlgn} illustrate the architecture of the {\abbgone} Network trained on Child and Alarm data, respectively. We use 20 noise variables as input to the {\abbgone} for Child data. The input to the  {\abbgone} trained on Alarm data consists of 37 noise variables.
\begin{table}[H]
\centering
\caption{{\abbgone} Network Architecture for Child dataset} 

\begin{tabular}{@{\extracolsep{4pt}}cc}
\toprule   
 \textbf{Discriminator} D &  \textbf{Generator} G \\
 \cmidrule{0-0}
 \cmidrule{2-2} 
$\text{Input} \in \mathbb{R}^{60}$ & $\text{Input} \in \mathbb{R}^{20}$\\ 
 \midrule
FC-(N100), LReLU(0.2) & FC-(N100), ReLU \\ 
 \midrule
FC-(N100), LReLU(0.2) & FC-(N100), BN, ReLU \\ 
 \midrule
FC-(N100), LReLU(0.2) & FC-(N100), BN, ReLU\\ 
 \midrule
FC-(N100), LReLU(0.2)& FC-(N100), BN, ReLU \\ 
 \midrule
\multirow{5}{*}{FC-(N100), LReLU(0.2)} &  $8 \times$ FC-(N2)\\ 
&$8 \times$ FC-(N3)\\
& $1 \times$ FC-(N4)\\
&$2 \times$ FC-(N5)\\
& $1 \times$ FC-(N6)\\
 \midrule
FC-(N1) & $20 \times$ Softmax\\ 

\bottomrule
\end{tabular}

\label{table:childlgn}
\end{table}

\begin{table}[H]
\centering
\caption{{\abbgone} Network Architecture for Alarm dataset} 

\begin{tabular}{@{\extracolsep{4pt}}cc}
\toprule   
 \textbf{Discriminator} D &  \textbf{Generator} G \\
 \cmidrule{0-0}
 \cmidrule{2-2} 
$\text{Input} \in \mathbb{R}^{105}$ & $\text{Input} \in \mathbb{R}^{37}$\\ 
 \midrule
FC-(N100), LReLU(0.2) & FC-(N100), ReLU \\ 
 \midrule
FC-(N100), LReLU(0.2) & FC-(N100), BN, ReLU \\ 
 \midrule
FC-(N100), LReLU(0.2) & FC-(N100), BN, ReLU\\ 
 \midrule
FC-(N100), LReLU(0.2)& FC-(N100), BN, ReLU \\ 
 \midrule
 \multirow{3}{*}{FC-(N100), LReLU(0.2)} &  $13 \times$ FC-(N2)\\ 
&$17 \times$ FC-(N3)\\
& $7 \times$ FC-(N4)\\
 \midrule
FC-(N1) & $37 \times$ Softmax\\
\bottomrule
\end{tabular}

\label{table:alarmlgn}
\end{table}




\subsection*{A.2.4. Baselines for Validating the Causal Graph Learned by the Model Experiment}
In the following, we discuss the implementation details of the baselines used in the ``Causal Graph Learned by the Model" experiment. All baselines are trained using the default hyperparameters unless otherwise stated.

\textbf{DAG-GNN \cite{yu2019dag}:} A causal discovery framework based on variational autoencoder, in which both decoder and encoder are graph neural networks. The weighted adjacency matrix of the causal graph is learned by optimizing the evidence lower bound. In our experiments, we use python implementation of the framework, available at first author's github repository \href{https://github.com/fishmoon1234/DAG-GNN}{https://github.com/fishmoon1234/DAG-GNN}.

\textbf{RL-BIC \cite{zhu2019causal}:} A causal discovery framework which leverages reinforcement learning to search the space of DAGs and find the optimal one. In our experiment, we use the python implementation by the first author of the paper, available at \href{https://github.com/huawei-noah/trustworthyAI}{https://github.com/huawei-noah/trustworthyAI}.

\textbf{PC-algorithm \cite{spirtes2000causation}:} A constraint-based method, which finds the skeleton of the graph using conditional independence test and orients the edges up to the Markov equivalence class. We use the python implementation by Causal Discovery ToolBox package \cite{kalainathan2019causal}.

\textbf{GSF \cite{huang2018generalized}:} A score-based causal discovery framework that adopts GES search strategy with a generalized score function which enables the model to account for nonlinear relationships. The Matlab implementation of the algorithm can be found at first author's github repository \href{https://github.com/Biwei-Huang/Generalized-Score-Functions-for-Causal-Discovery}{https://github.com/Biwei-Huang/Generalized-Score-Functions-for-Causal-Discovery}.

\section*{A.3. Additional Experimental Results}

\begin{figure*}%
    \centering
    \subfloat[]{{\includegraphics[width=8cm, height = 3.7cm]{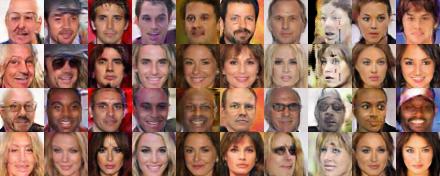} }}\hspace{0.7mm}%
    \qquad
    \subfloat[]{{\includegraphics[width=8cm, height = 3.7cm]{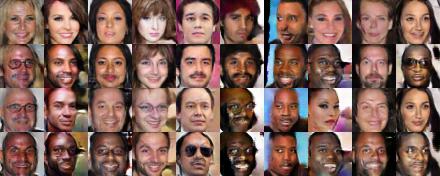} }}%
    \caption{Results of {\m} for intervening and conditioning on labels : (a) shows the results for "Wearing-Lipstick" label. Since in the causal graph learned (Figure \ref{fig:causalgraph}) we have $Male \rightarrow Wearing-Lipstick$, we do not expect intervening on Wearing-Lipstick to affect the distribution of Male, i.e., $\mathbb{P}(\text{Male}=1|\text{do(Wearing-Lipstick=1)})=\mathbb{P}(\text{Male}=1)=0.41$. Thus, in the interventional samples we observe both males and females wearing lipstick. On the other hand, conditioning on Wearing-Lipstick affects the distribution of Male, i.e., $\mathbb{P}(\text{Male}=1|\text{Wearing-Lipstick=1})\approx0$. Hence, in the conditional samples, only females with lipstick can be found. (b) shows the results for "Mustache" label. Since in the learned causal graph we have $Wearing-Lipstick \rightarrow Mustache$, intervening on Mustache should not affect the distribution of Wearing-Lipstick, i.e., $\mathbb{P}(\text{Wearing-Lipstick}=1|\text{do(Mustache=1)})=\mathbb{P}(\text{Wearing-Lipstick}=1)=0.47$. This explains the existence of people with mustache both with and without lipstick in the interventional samples. Note that conditioning on Mustache affects the distribution of Wearing-Lipstick, i.e., $\mathbb{P}(\text{Wearing-Lipstick}=1|\text{Mustache=1})\approx0$. Hence, no person with mustache is seen wearing a lipstick in the conditional samples.}%
    \label{fig:lips}%
\end{figure*}

\begin{figure*}%
    \centering
    \subfloat[]{{\includegraphics[width=8cm, height = 3.7cm]{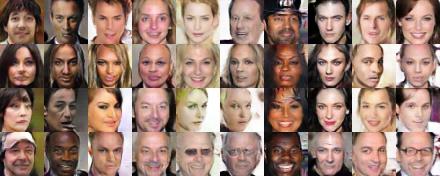} }}\hspace{0.7mm}%
    \qquad
    \subfloat[]{{\includegraphics[width=8cm, height = 3.7cm]{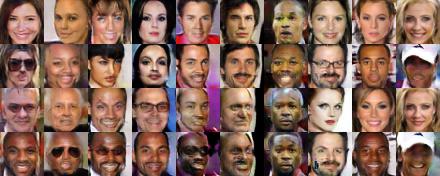} }}%
    \caption{Results of {\m} for intervening and conditioning on labels : (a) shows the results for "Bald" label. According to the causal graph (Figure \ref{fig:causalgraph}), we have $Young \rightarrow Bald$, therefore, intervening on Bald does not affect Young, i.e., $\mathbb{P}(\text{Young}=1|\text{do(Bald=1)})=\mathbb{P}(\text{Young}=1)=0.77$. However, conditioning on Bald changes the  $\mathbb{P}(\text{Young}=1)$ from 0.77 to 0.23.
     (b) shows the results for "Mustache" label. Since in the learned causal graph we have $Male\rightarrow Mustache$, intervening on Mustache should not affect the distribution of Male, i.e., $\mathbb{P}(\text{Male}=1|\text{do(Mustache=1)})=\mathbb{P}(\text{Male}=1)=0.41$. However, we have $\mathbb{P}(\text{Male}=1|\text{Mustache=1})\approx1)$, which means the conditional samples should only contain males whereas in the interventional samples we have both males and females with mustache.}%
    \label{fig:young}%
\end{figure*}

\subsection*{A.3.1. Additional Qualitative Results}

In this section, we present additional images generated by {\m} from both conditional and interventional distributions in Figures \ref{fig:lips} and \ref{fig:young}. Note that in these figures, top two rows show the results for intervention (i.e., intervention on label=0 and label=1, respectively) and bottom two rows are the results for conditioning (i.e.,  condition on label=0 and label=1, respectively).

\begin{table*}[!htbp]
  \begin{center}
  \caption{The GAN-train and GAN-test scores for {\m} and baselines for all 9 labels in CelebA separately. Analyzing GAN-train and GAN-test (higher is better) demonstrates the effectiveness of {\m}}
    \label{table1:accuracy}
    \begin{tabular}{@{}cccc@{}} \toprule

Label& Model  & GAN-train & GAN-test\\ \midrule
\multirow{3}{*}{Bald}&{\m}& 97.72& 87.67\\
&CausalGAN & 91.11 & 59.43\\
&AC-GAN  & 91.15& 57.96\\
\midrule
\multirow{3}{*}{Eyeglasses}&{\m}& 96.01& 91.62\\
&CausalGAN & 92.31 & 87.96\\
&AC-GAN  & 95.24& 87.05\\
\midrule
\multirow{3}{*}{Male}&{\m}& 86.56& 88.12\\
&CausalGAN & 85.24 & 76.40\\
&AC-GAN  & 87.00& 70.70\\
\midrule
\multirow{3}{*}{\text{Mouth-Slightly-Open}}&{\m}& 79.26& 80.00\\
&CausalGAN & 75.43 & 72.36\\
&AC-GAN  & 79.41& 70.85\\
\midrule
\multirow{3}{*}{Mustache}&{\m}& 90.15& 90.74\\
&CausalGAN & 82.66 & 59.32\\
&AC-GAN  & 88.58& 57.98\\
\midrule
\multirow{3}{*}{Narrow-Eyes}&{\m}& 82.33& 82.90\\
&CausalGAN & 61.14 & 50.23\\
&AC-GAN  & 69.01& 52.51\\
\midrule
\multirow{3}{*}{Smiling}&{\m}& 82.86& 86.28\\
&CausalGAN & 71.83 & 72.41\\
&AC-GAN  & 78.55& 71.26\\
\midrule
\multirow{3}{*}{Wearing-Lipstick}&{\m}& 80.07& 85.50\\
&CausalGAN & 72.30 & 59.96\\
&AC-GAN  & 71.09& 55.52\\
\midrule
\multirow{3}{*}{Young}&{\m}& 67.37& 64.34\\
&CausalGAN & 65.93 & 63.14\\
&AC-GAN  & 76.39& 61.59\\
\bottomrule
\end{tabular}
  \end{center}
  
\end{table*}

\begin{table*}[!htbp]
\centering
\caption{Evaluation of the quality of labels on Alarm dataset} 
\begin{tabular}{@{\extracolsep{4pt}}cccc}
\toprule   
 {} & \multicolumn{3}{c}{CelebA-label} \\
 \cmidrule{2-4} 

 Model  &  $\text{MSE}_p$ & $\text{MSE}_f$ & $\text{MSE}_a$ \\ 
\midrule
{\abbgone}-{\m} & \num{4.1e-4} $\pm$ \num{2e-5}  & \num{4.9e-3} $\pm$ \num{2e-3} & \num{3.3e-4} $\pm$ \num{2e-5} \\ 
 
MC-WGAN-GP& \num{4.8e-4} $\pm$ \num{5e-5}& \num{7.1e-3} $\pm$ \num{2e-3}& \num{6.3e-4} $\pm$ \num{7e-5}\\ 

Causal Controller & - & - & - \\ 
\bottomrule
\end{tabular}

\label{table:alarmlabel}
\end{table*}
\begin{figure}
\centering
\includegraphics[width=80mm,scale=1.4]{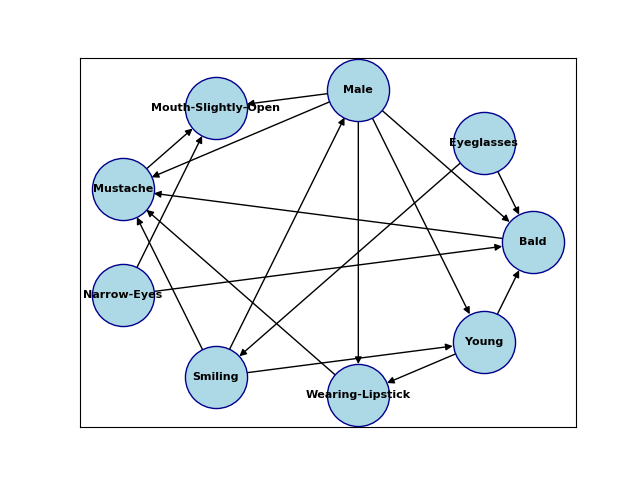}
\caption{Causal graph over the 9 labels of CelebA data. This graph is learned with {\abbgone} model from the data.} 
\label{fig:causalgraph}
\end{figure}

\subsection*{A.3.2. Quantitative Evaluation: Evaluation Metrics Details}
\textbf{GAN-train and GAN-test:}
In this section, we explain the details of the two metrics, namely GAN-train and GAN-test, which are used to evaluate the proposed {\abbgtwo} network. More specifically, to calculate GAN-train, a classification network is trained with images generated by the GAN model,
and then its performance is evaluated on a test set  of real-world images. To calculate the GAN-test, on the other hand, a classification network is trained on the real-world data and tested on the images generated by the GAN. In our experiments, for the classification network, we train Resnet-18 \cite{he2016deep} with learning rate 0.001 for 20 epochs. We also extend the original architecture to perform multi-label classification. Since our results are for multiple labels, we report both the hamming score, a label-based accuracy metric for multi-label classification (Table (1) in the main text) as well as classification accuracy for each label. The classification accuracy per each label is summarized in table \ref{table1:accuracy}. The results show that our proposed {\m} outperforms both baselines, i.e. CausalGAN and AC-GAN and demonstrate the effectiveness of our model in generating high-quality and diverse images compared to the baselines which do not consider the causal relations amongst pixels and labels. 
\subsection*{A.3.3. Label Generation Evaluation: Alarm Data Results}
Results for the evaluation of samples generated for Alarm data by {\abbgone} are shown is table \ref{table:alarmlabel}. The results show that {\abbgone} outperforms MC-WGAN-GP which does not consider the causal relations between features of a sample. Note that since Causal Controller component of CausalGAN cannot be trained on Alarm dataset due to scalability issue, we do not report the results for Causal Controller.

\subsection*{A.4. Causal Graph for CelebA} \label{causalgraph}
In this section, we show the causal graph learned by the {\abbgone} on 9 labels of the CelebA data in Figure \ref{fig:causalgraph}. Particularly, this graph
is learned by the {\abbgone} trained on the CelebA-Label data. Some of the learned relationships are as follows: $Male \rightarrow Mustache$, $Male \rightarrow Bald$, $Male \rightarrow Wearing-Lipstick$ and $Young \rightarrow Bald$ which are in compliance with our expectations.

\bibliographystyle{unsrt}
\bibliography{egbib}

\end{document}